\documentclass[times,twocolumn,final,authoryear]{elsarticle}
\pdfoutput=1
\usepackage{prletters}
\usepackage{framed,multirow}

\usepackage{amssymb}
\usepackage{latexsym}

\usepackage{amsmath}
\usepackage{bm}
\usepackage{float}
\usepackage{subfig}
\usepackage{caption}
\usepackage{array}
\usepackage{footnote}
\usepackage{footmisc}

\renewcommand{\thefootnote}{\alph{footnote}}
\newcommand{\astfootnote}[1]{
\let\oldthefootnote=\thefootnote
\setcounter{footnote}{0}
\renewcommand{\thefootnote}{\fnsymbol{footnote}}
\footnote{#1}
\let\thefootnote=\oldthefootnote
}

\newcolumntype{x}[1]{>{\centering\arraybackslash\hspace{0pt}}p{#1}}
\usepackage{url}
\usepackage[dvipsnames]{xcolor}
\definecolor{newcolor}{rgb}{.8,.349,.1}

\journal{Pattern Recognition Letters}

\begin{document}

\thispagestyle{empty}

\ifpreprint
  \setcounter{page}{1}
\else
  \setcounter{page}{1}
\fi

\begin{frontmatter}

\title{Mean Oriented Riesz Features for Micro Expression Classification}

\author{Carlos \snm{Arango Duque}} 
\author{Olivier \snm{Alata}\corref{cor1}}
\cortext[cor1]{Corresponding author: 
  Tel.: +33(0)469663262;}
\ead{olivier.alata@univ-st-etienne.fr}
\author{R\'{e}mi \snm{Emonet}}
\author{Hubert \snm{Konik}}
\author{Anne-Claire \snm{Legrand}}

%\author[1]{Carlos \snm{Arango Duque}\corref{cor1}} 
%\cortext[cor1]{Corresponding author: 
%  Tel.: +0-000-000-0000;}
%\ead{author@author.com}
%\author[2]{Olivier \snm{Alata}}
%\author[2]{R\'{e}mi \snm{Emonet}}
%\author[2]{Anne-Claire \snm{Legrand}}
%\author[2]{Hubert \snm{Konik}}

%\address[1]{Affiliation 1, Address, City and Postal Code, Country}
\address{Lab. Hubert Curien, CNRS UMR 5516, UJM, IOGS, Univ. Lyon, 42023 Saint Etienne, France}

\received{1 May 2013}
\finalform{10 May 2013}
\accepted{13 May 2013}
\availableonline{15 May 2013}
\communicated{S. Sarkar}

\begin{abstract}
Micro-expressions are brief and subtle facial expressions that go on and off the face in a fraction of a second. This kind of facial expressions usually occurs in high stake situations and is considered to reflect a human's real intent. There has been some interest in micro-expression analysis, however, a great majority of the methods are based on classically established computer vision methods such as local binary patterns, histogram of gradients and optical flow. A novel methodology for micro-expression recognition using the Riesz pyramid, a multi-scale steerable Hilbert transform is presented. In fact, an image sequence is transformed with this tool, then the image phase variations are extracted and filtered as proxies for motion. Furthermore, the dominant orientation constancy from the Riesz transform is exploited to average the micro-expression sequence into an image pair. Based on that, the Mean Oriented Riesz Feature description is introduced. Finally the performance of our methods are tested in two spontaneous micro-expressions databases and compared to state-of-the-art methods.
\end{abstract}

\begin{keyword}
\MSC 68U10\sep 68T10
\KWD Micro-expressions\sep Mean Oriented Riez Features\sep Recognition\sep Classification\sep Subtle motion analysis\sep Riesz pyramid\sep Monogenic signal\sep SMIC database\sep CASME2 database
%% MSC codes here, in the form: \MSC code \sep code
%% or \MSC[2008] code \sep code (2000 is the default)
\end{keyword}

\end{frontmatter}

%\linenumbers

%% main text
\section{Introduction}
\label{sec1}

Micro-expressions (MEs) are brief and subtle facial expressions that last a fraction of a second which are considered to reflect a human's hidden emotions \citep{ekman_smiles_2005}. Analyzing them has become a challenging problem in computer vision with different state-of-the-art approaches based on well-known computer vision methods such as local binary patterns (LBP), histogram of gradients (HOG) and optical flow (OF). However, MEs are composed of subtle motions which might be difficult to process with classical approaches. One possible solution is to analyze them by comparing the phase variations between images. \citep{fleet_computation_1990,gautama_phase-based_2002} initially proposed that spatio-temporally band-passed video provides a good approximation to the motion field.

However, its true potential became evident with motion magnification a method in which phase variations of subtle motions are amplified~\citep{wadhwa_phase-based_2013}. Specifically, the Riesz pyramid-based representation for video magnification \citep{wadhwa_riesz_2014} has shown to be a simple, adaptable and fast-processing method that can work in almost real-time. In addition to allowing the motion to be exaggerated, the intermediate representations produced by these methods can be used to directly analyze subtle motion. Indeed, it has been already used for subtle motion analysis~\citep{visapp18} and ME spotting~\citep{8354118}.

This paper proposes a framework based on this tool to extract multi-scale oriented phase variation features using the Riesz pyramid in order to model and classify MEs. There have been some other authors who have proposed to extract phase variations using a Riesz wavelet or transform as part of their ME classification scheme \citep{oh_monogenic_2015,oh_intrinsic_2016,liong_micro-expression_2017}. However, to our knowledge, there is no other method that uses the multi-scale phase variations as the main feature of their solution.

This paper is divided as follows: Sec.~\ref{sec:state} presents a brief recapitulation of the state of the art in ME recognition. Sec.~\ref{sec:mono} introduces the reader to the monogenic signal components and the Riesz pyramid.
Sec.~\ref{sec:MORF} presents a novel way to extract motion features using the orientation and phase from the monogenic signal to model MEs. Sec.~\ref{sec:exp} analyzes the results of our experiments and compares them with state-of-the-art methods. Finally, in Sec.~\ref{sec:conc}, our conclusions are presented.

\section{Related Work}
\label{sec:state}
The ME recognition frameworks can be divided into different feature representation families. The first one is composed of \textit{LBP-based methods}. They use the intensity information of the image with the intention of describing facial features that appear temporally during any kind of facial expression. A great numbers of descriptors are based on local binary patterns (LBP) introduced in \citep{zhao_dynamic_2007}. A 3D extension called LBP from Three Orthogonal Planes (LBP-TOP) takes a stack of consecutive frames as 3D volume and compute LBP over three orthogonal planes and has been used in several ME recognition frameworks~\citep{li_spontaneous_2013}.
% ~\citep{li_towards_2017,ngo_eulerian_2016,park_subtle_2015,pfister_recognising_2011,li_micro-expression_2019}. 
%\citep{wang_micro-expression_2015} projects the images into a tensor independent color space (TICS) before applying LBP-TOP. 
Due to the popularity of LBP-TOP, a plethora of variations have emerged for feature extraction. 
% For instance, LBP with Six Intersection Points (LBP-SIP) that reduces the redundancy in LBP-TOP patterns\citep{wang_lbp_2014}. 
For instance, some authors propose to take the average plane from each stack first, and then compute the LBP on the three average planes (MOP-LBP)~\citep{wang_efficient_2015}. Another variation called Spatio-Temporal LBP with Improved Integral Projection (STLBP-IIP) preserves the shape property of MEs and then enhances discrimination of the features for ME recognition~\citep{huang_facial_nodate,huang_spontaneous_2016}. Another proposal called Spatio-temporal Completed Local Quantized Pattern (STCLQP) extracts sign, magnitude and orientation information while creating a compact and discriminative codebook~\citep{huang_spontaneous_2016-1}. 
% Some authors have proposed a two-step approach by first extracting the sparse part of Robust PCA (RPCA) of an image and then computing Local Spatio-Temporal Directional features, providing a (RPCA-LSTD) descriptor which is analogous to LBP-TOP~\citep{wang_micro-expression_2014}.

The second family is composed of \textit{OF-based methods}. They use the distribution of the apparent velocities of objects in an image for motion representation. Some authors propose to derive the OF vectors to calculate the optical strain (OS) or non-rigid deformation for the analysis of MEs~\citep{liong_optical_2014}. 
Inspired in the success of histogram features in the object recognition community, \citep{chaudhry_histograms_2009} proposed the Histogram of Oriented OF (HOOF) descriptor to model the distribution of OF during a video sequence. Another approach called the fuzzy histogram of oriented optical flow orientations (FHOFO) collects the motion directions into angular bins based on the fuzzy membership function~\citep{happy_fuzzy_2018}. Some similar descriptors like Main Directional Mean Optical flow (MDMO)~\citep{liu_main_2016} and Facial Dynamic Maps (FDM)~\citep{xu_microexpression_2016} extract OF motion vectors from selected facial regions. 
% Another descriptor called Fusion Motion Boundary Histograms (FMBH) creates a series of weighted histograms from the horizontal and vertical derivatives of OF~\citep{lu_motion_2018}.
Other authors have proposed to calculate which facial regions have high probabilities of movement (RHPM) and use them to filter the OF~\citep{allaert_consistent_2017}. 
Another approach called Bi-Weighted oriented OF (BI-WOOF) is a variation of HOOF that uses optical strain as a weighting coefficient~\citep{liong_less_2018}. 

The third family is composed of \textit{Deep learning methods}. These methods normally combine feature learning and classification in the same pipeline.
% In this one Learned features are usually trained through a joint feature learning and classification pipeline~\citep{corneanu_survey_2016}. 
% For instance, \citep{he_multi-task_2017} proposes a multi-task mid-level feature learning method to enhance the discrimination ability of extracted low-level features by learning a set of class-specific feature mappings, which would be used for generating a mid-level feature representation. 
In~\citep{kim_micro-expression_2016}, the spatial features of micro-expressions at different expression-states are encoded using a CNN and then the temporal characteristics of the different expression-states of the micro-expressions are encoded using long short-term memory (LSTM) recurrent neural networks. In~\citep{li_can_2018}, a VGGNet trained for face recognition is finely tuned and adapted for ME recognition. Other authors propose to extract the ME features using a pre-trained network (ImageNET) and an evolutionary feature selection scheme to remove the irrelevant deep features \citep{patel_selective_2016}. In~\citep{li_micro-expression_2018}, a CNN is trained using both gray-scale images and optical flow as input (3D-FCNN).

There are certain proposals that do not really fit the previous families of methods.
For instance, \citep{lin_micro-expression_2018} uses spatio-temporal Gabor filters (ST-Gabor) to extract features at different scales and orientations by convolving a bank of oriented bandpass filters to an image sequence. 
% Another approach, \citep{oh_monogenic_2015}, uses the Riesz wavelet to transform the input images into multi-scale monogenic wavelets, then it extracts features from the magnitude, orientation and phase from each scale. 
Some approaches also consider combining appearance-based features and OF. For instance, \citep{liong_micro-expression_2017} proposes to mix the BI-WOOF and the phase components from the monogenic signal obtained by a Riesz transform. Furthermore, \citep{liong_subtle_2014} uses optical strain as weights for LBP-TOP (OSW-LBP-TOP).

In this paper, a complete framework for ME recognition based on the Riesz pyramid representation, a fast multi-scale approximation of the Riesz transform is presented. A new feature called Mean Oriented Riesz Feature (MORF), using the multi-scale oriented phase of the Riesz pyramid is introduced. Let us now recall the basis of the Riesz pyramid and its components.

\begin{figure*}[h]
	\centering
    \includegraphics[width=0.9\textwidth]{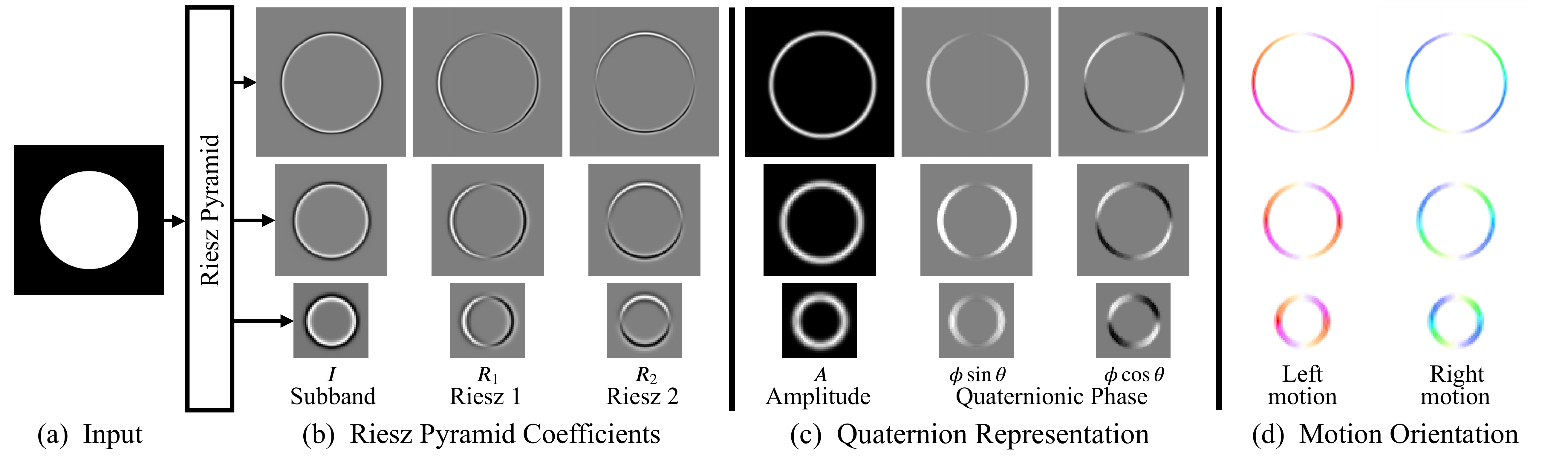}
    % \vspace*{-10mm}
	\caption{Different representations of the Riesz pyramid. The input is a circle with a sharp edge (a). In (b), the input is decomposed into multiple spatial sub-bands using an invertible transform, and an approximate Riesz transform is taken of each band to form the Riesz pyramid. In (c), the Riesz pyramid coefficients are transformed into a quaternion. Then, for each subband the amplitude and the quaternionic phase can be extracted. In (d), we show the quaternionic phase difference between two consecutive frames for the input image translated one pixel to the left (Left motion) and one pixel to the right (Right motion).}
	\label{fig:Riesz_pyr}
\end{figure*}

\section{Background}
\label{sec:mono}
\subsection{The Monogenic Signal and the Riesz Transform}
In signal analysis, a real valued 1-D signal can be represented as a complex valued signal. From this representation some useful information can be extracted such as the local amplitude and local phase. The analytical representation is composed of the original function and its Hilbert transform. In the case of 2-D signals (images), \citep{felsberg_monogenic_2001} proposed an isotropic generalization called the monogenic signal. The monogenic signal is a triple comprised of the original image and a quadrature pair produced by the Riesz transform (a 2-D steerable generalization of the Hilbert transform). This quadrature pair is 90 degrees phase-shifted with respect to the dominant orientation at every pixel~\citep{wadhwa_riesz_2014}, thus we can extract the local amplitude and the local phase variations in the direction of the dominant orientation from the monogenic signal.
Let $I(\boldsymbol{x})$ be a 2D gray scale image of a spatial variable $\boldsymbol{x} = (x,y)^\intercal$, and let $F(\boldsymbol{\omega})$ be its frequency-domain representation found using the 2D Fourier transform, where $\boldsymbol{\omega} = (\omega_1, \omega_2)^\intercal$ is a two-dimensional frequency~\citep{bridge_introduction_2017}. The two odd parts $F_{R_1}$ and $F_{R_2}$ of the monogenic signal are:
\begin{align}
	\begin{split}
	F_{R_l}(\boldsymbol{\omega}) & = \Bigg\{
	\begin{array}{ll}
		i\frac{\omega_{l} }{\|\boldsymbol{\omega}\|}F(\boldsymbol{\omega}), & \boldsymbol{\omega} \neq 0 			\\ 0, & \boldsymbol{\omega} = 0 \\
	\end{array} \\ 
	\end{split}
	\label{eq:rieszfreq}
\end{align}
where $l = 1$ or $2$ and $R_1(\boldsymbol{x})$ and $R_2(\boldsymbol{x})$ correspond to the image domain representation of $F_{R_1}$ and $F_{R_2}$ respectively.
\subsection{The Riesz pyramid}
The Riesz pyramid decomposes the image into multiple sub-bands, each of which corresponds to a different spatial scale, and then applies the Riesz transform of each sub-band (Fig.\ref{fig:Riesz_pyr}b). An ideal version of the Riesz pyramid could be built in the frequency domain using octave (or sub-octave) filters similar to the ones proposed in \citep{wadhwa_phase-based_2013} and the frequency domain Riesz transform (Eq.~\ref{eq:rieszfreq}). However, it requires the use of costly Fourier transforms to be built. In order to make the Riesz pyramid faster, some adaptations need to be made. Firstly, instead of using the Fourier transform, the image is decomposed into non-oriented sub-bands using an invertible image pyramid such as the Laplacian pyramid. Secondly, an approximate Riesz transform can be defined by two finite difference filters, which is significantly more efficient to compute. Since most of the energy from the previously processed sub-bands are concentrated in a frequency band around $\|\boldsymbol{\omega}\| = \frac{\pi}{2}$, the Riesz transform can be approximated with the three tap finite difference filters $[0.5, 0, -0.5]$ and $[0.5, 0, -0.5]^\intercal$~\citep{wadhwa_riesz_2014}.

\subsection{Riesz pyramid Coefficients}
\label{sec:rieszcoef}
If a given image subband $I$ is filtered using this method, the result is the pair of filter responses, $(R_1;R_2)$. The input $I$ and the Riesz transform $(R_1;R_2)$ together form a triple (the monogenic signal) that can be converted to spherical coordinates (applying the method from ~\citep{wadhwa_riesz_2014}) to yield the local amplitude $A$, local orientation $\theta$ and local phase $\phi$ from the Riesz coefficients.
\begin{equation}
	\begin{split}
		I & = A\cos{(\phi)} \\ 
		R_1 & = A\sin{(\phi)}\cos{(\theta)} \\
		R_2 & = A\sin{(\phi)}\sin{(\theta)}
	\end{split}	
	\label{eq:rieszmono}
\end{equation}
While Eq.~\ref{eq:rieszmono} can be solved, both $(A,\phi,\theta)$ and $(A,-\phi,\theta + \pi)$ are possible solutions. This predicament can be fixed by considering
\begin{equation}
	\phi\cos{(\theta)}, \phi\sin{(\theta)}
	\label{eq:phicos}
\end{equation}	 
which are invariant to this sign ambiguity. If the methods of ~\citep{wadhwa_quaternionic_2014} in which the Riesz pyramid coefficients are represented as a quaternion are applied:
 
\begin{equation}
	\textbf{r} = I + iR_1 + jR_2
\end{equation}
then, the previous equation can be rewritten using Eq.~\ref{eq:rieszmono} as:
\begin{equation}
	\textbf{r} = A\cos{(\phi)} + iA\sin{(\phi)}\cos{(\theta)} + jA\sin{(\phi)}\sin{(\theta)}
	\label{eq:riesz}
\end{equation} 
Thus, the local amplitude $A$ and the quaternionic phase $(\phi\cos{(\theta)}, \phi\sin{(\theta)})$ are computed as: 
\begin{equation}
	\begin{aligned}
		A & = \|\textbf{r}\| \\
		i\phi\cos{(\theta)} + j\phi\sin{(\theta)} & = \log{(\textbf{r}/\|\textbf{r}\|)}
	\end{aligned}
	\label{eq:quatlog}
\end{equation}
Furthermore, the quaternionic phase can be denoised by applying a temporal quaternionic filtering scheme. A complete explanation of the quaternionic operation used to extract the Riesz coefficients and filtering can be found in \citep{wadhwa_quaternionic_2014}. Fig.\ref{fig:Riesz_pyr}c shows the local amplitude and filtered quaternionic phase extracted from different levels of the Riesz pyramid applied to an input image translated one pixel to the left.

\subsection{Local orientation of the quaternionic phase}
\label{sec:locorien}
One of the advantages of the monogenic signal is that, in the same way as the analytical signal, it preserves the split of identity. This means that, the local phase is invariant to changes of the local orientation, and the local orientation is invariant to changes of the local structure (which means that we can split them). If we can recover the correct local direction from the local orientation, we have an ideal split of identity with respect to energetic, geometric, and structural information of the signal. However, there are a few problems in estimating the correct local direction. The first one is that the estimation of local orientation is unstable if the local phase $\phi$ is close to $0$. The second problem is that it's not possible to find an absolute estimation for the local direction but rather the relative estimation. 

The main question becomes whether the orientation component of the quaternionic phase can be used to differentiate between opposing motions. Fig.\ref{fig:Riesz_pyr}d presents a circle with a sharp edge which is translated one pixel in any given direction. The resulting filtered quaternionic phase with pseudo-colors where the image saturation represents the phase $\phi$ component and the color hue represents the orientation $\theta$ component is presented. The areas of low amplitude are masked using the technique presented in \citep{visapp18} for better visualization. The image is translated one pixel to the left (left motion) and to the right (right motion).In that way motion in different directions can be represented (as evidenced by the different hues from the pseudo color image representations from the edges of  Fig.\ref{fig:Riesz_pyr}d). The orientation of the quaternionic phase is not the same as that of the translation but rather it is perpendicular to the orientation of the edge. This means that the oriented quaternionic phase is affected by the aperture problem. Nevertheless, when comparing two opposing motions (left vs right), their respective oriented quaternionic phases are also opposite. This becomes important for ME recognition, where different MEs represent motions in different directions. For example, when analysing the eyebrow movement during an ME of surprise, the eyebrows rise. On the other hand during an ME of anger, the eyebrows are contracted (lowered).

\begin{figure*}[]
	\centering
    \includegraphics[width=0.9\textwidth]{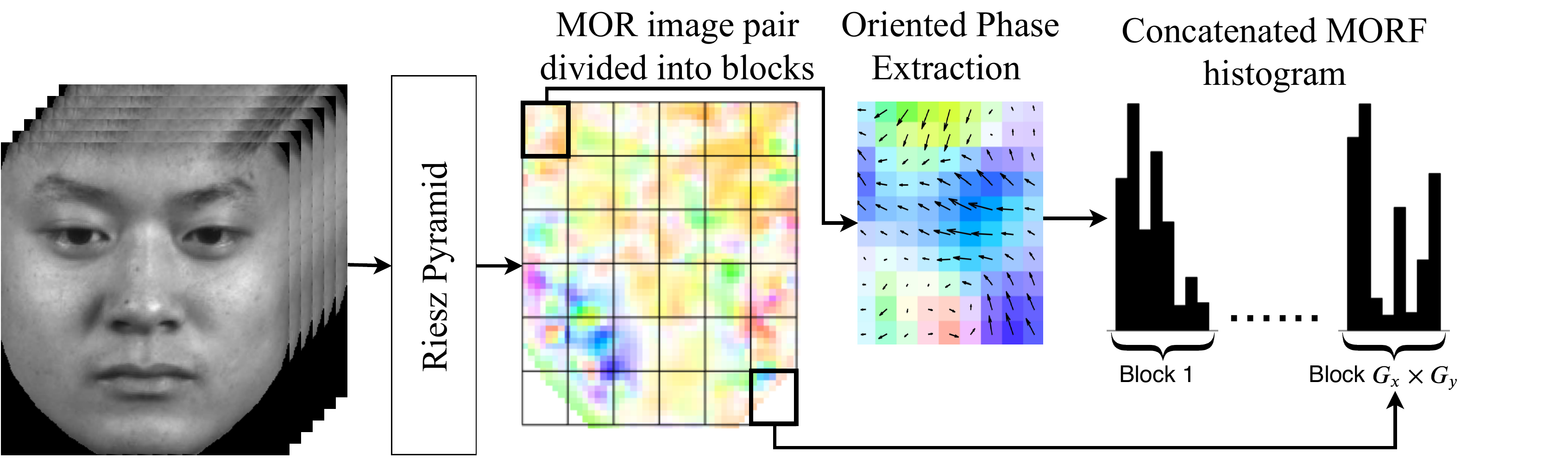}
    % \vspace*{-10mm}
	\caption{Extraction of the MORF descriptor}
	\label{fig:MORF_flow3}
\end{figure*}

\section{Mean Oriented Riesz Features}
\label{sec:MORF}
This paper now proposes a descriptor to extract the oriented phase elements from the monogenic signal called \textbf{Mean Oriented Riesz Features (MORF)}. While Sec.~\ref{sec:meandirpair} introduces the concept of the mean oriented Riesz image pair, Sec.~\ref{sec:MORFimp} describes the implementation of our proposed descriptor.

\subsection{Mean Oriented Riesz Image Pair}
\label{sec:meandirpair}
In Sec.\ref{sec:locorien} it is shown how using a relative quaternion phase estimation motions from different directions can be differentiated. However, only the motion between two consecutive frames is analyzed. Considering the MEs are captured as video sequences of several frames, it is also necessary to analyse the temporal evolution of these motions before proposing an ME modelling scheme.
It can be considered unnecessary to analyse the whole ME sequence but rather a shorter sequence from ME onset to ME apex (when the face goes from a neutral state to a state of peak expressiveness) because the spatial displacement of the facial muscles is more evident compared to the sequence that goes from ME apex to ME offset (the face goes from peak expressiveness to a neutral state). It is also necessary to deal with the high variance of the quaternionic phase $(\phi\cos(\theta),\phi\sin(\theta))$ in areas of low local amplitude $A$. \citep{8354118} proposes to crop a series of local ROIs and to mask them using the local amplitude from the Riesz pyramid in order to isolate areas of potential noise. Although this approach was effective for ME spotting, it ignores certain facial areas of low amplitude which might have some interesting information while an ME is taking place (such as the cheek areas).

Taking the aforementioned considerations into account, we propose to model the temporal evolution of the ME in two single images called the \textbf{mean oriented Riesz (MOR) image pair}. The filtered quaternionic phase of an ME sequence is simply calculated from onset to apex and then, for each pixel, the results are averaged through the time axis : 
\begin{align}
	\overline{\phi\lambda{(\theta)}} &= \frac{1}{f_a-f_o+1}\sum_{t=f_o}^{f_a} \phi_{t}\lambda{(\theta_t)} 
	\label{eq:meanphicos}
\end{align}
with $f_o$ and $f_a$, the frame the onset begins and the frame of the apex respectively, and $\lambda$ either $\cos$ or $\sin$.
The main intuition is that by temporally averaging the filtered quaternionic phase, the real motion of each pixel is modelled in a single orientation and magnitude while reducing the effect of wrongfully detected motion due to noise. 

\begin{figure}[H]
    \centering
    \textbf{Compare Pyramid Level Effect}\par\medskip
    \includegraphics[width=0.6\columnwidth]{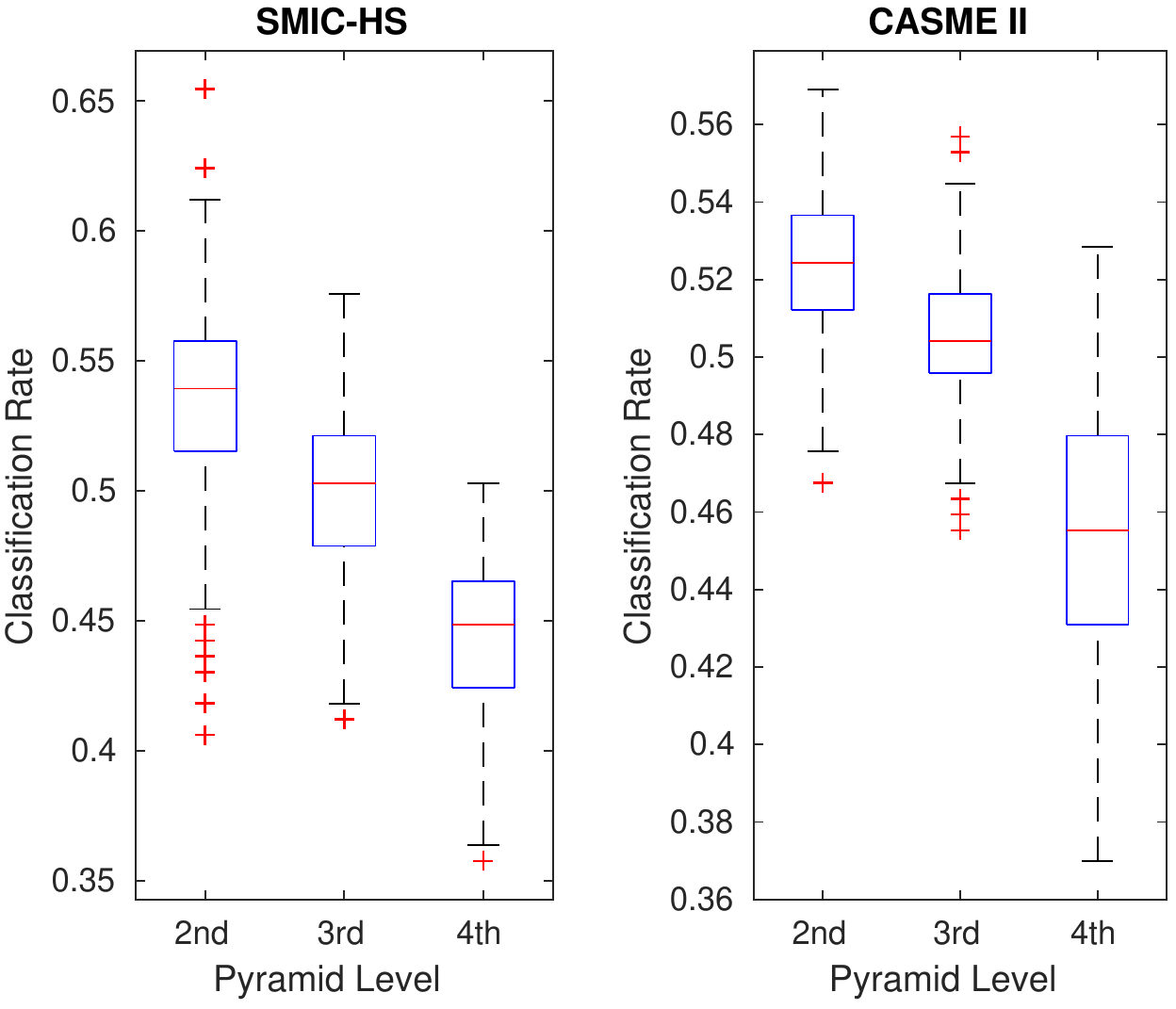}
    \caption{Riesz pyramid level evaluation}
    \label{fig:paralev}
\end{figure}

\subsection{MORF extraction}
\label{sec:MORFimp}
To begin with the face is detected in the first frame \citep{viola_rapid_2001}, then, an active appearance model  (AAM)~\citep{tzimiropoulos_optimization_2013} is used to detect a set of facial landmarks. Next, certain facial landmarks which will not move during facial expressions are selected (the inner corners of the eyes and the lower point of the nose between the nostrils are selected). These points are tracked using the KLT algorithm~\citep{tomasi_detection_1991} and a cropped face image sequence is obtained (the area outside the face border is masked - see Fig. \ref{fig:MORF_flow3}). The Riesz pyramid is applied to obtain the quaternionic phase (Sec.~\ref{sec:rieszcoef}) and it is filtered with the method  described in \citep{8354118}. Then, Eq.~\ref{eq:meanphicos} is then used to obtain the MOR image pair (see center of Fig. \ref{fig:MORF_flow3}). 

The face is divided into a grid of equally sized non-overlapping rectangle areas. As can be seen in Fig.~\ref{fig:MORF_flow3}, each pixel in the image pair represents a motion vector with a magnitude and angle. They can be extracted from the oriented phase by:
\begin{align}
	\overline{\phi_R} &= \sqrt[2]{(\overline{\phi_R\cos(\theta_R)})^2+(\overline{\phi_R\sin(\theta_R)})^2} \label{eq:histphi}\\
	\overline{\theta_R} &= \arctan\left(\frac{\overline{\phi_R\sin{(\theta_R)}}}{\overline{\phi_R\cos{(\theta_R)}}}\right)
	\label{eq:histtheta}
\end{align}
where $\overline{\phi_R}$ is a matrix containing the phase of every pixel, $\overline{\theta_R}$ is a matrix containing the dominant orientation of every pixel and $R$ corresponds to the level of the Riesz pyramid, the oriented phase is being extracted from.

we are extracting the oriented phase from. The next step is to create the histogram of oriented phase for each one of the rectangular blocks. For each pixel, a bin is selected based on the orientation $\theta$ and a weighted vote is cast based on the value of the phase $\phi$. The final histogram is the concatenation of all the histograms (Fig.~\ref{fig:MORF_flow3}).

The MORF descriptor depends on three parameters: $G$ which determines the grid division of ($[G_x,G_y]$) ROIs, $O$ which determines the number of orientations bins of the descriptor and $R$ which determines the level of the Riesz pyramid going to be extracted. Thus $\text{MORF}_{G,O,R}$ produces a feature vector of $G_x \times G_y\times O$ length.

\begin{figure}[H]
    \centering
    \textbf{Compare Angle Division Effect}\par\medskip
    \includegraphics[width=0.6\columnwidth]{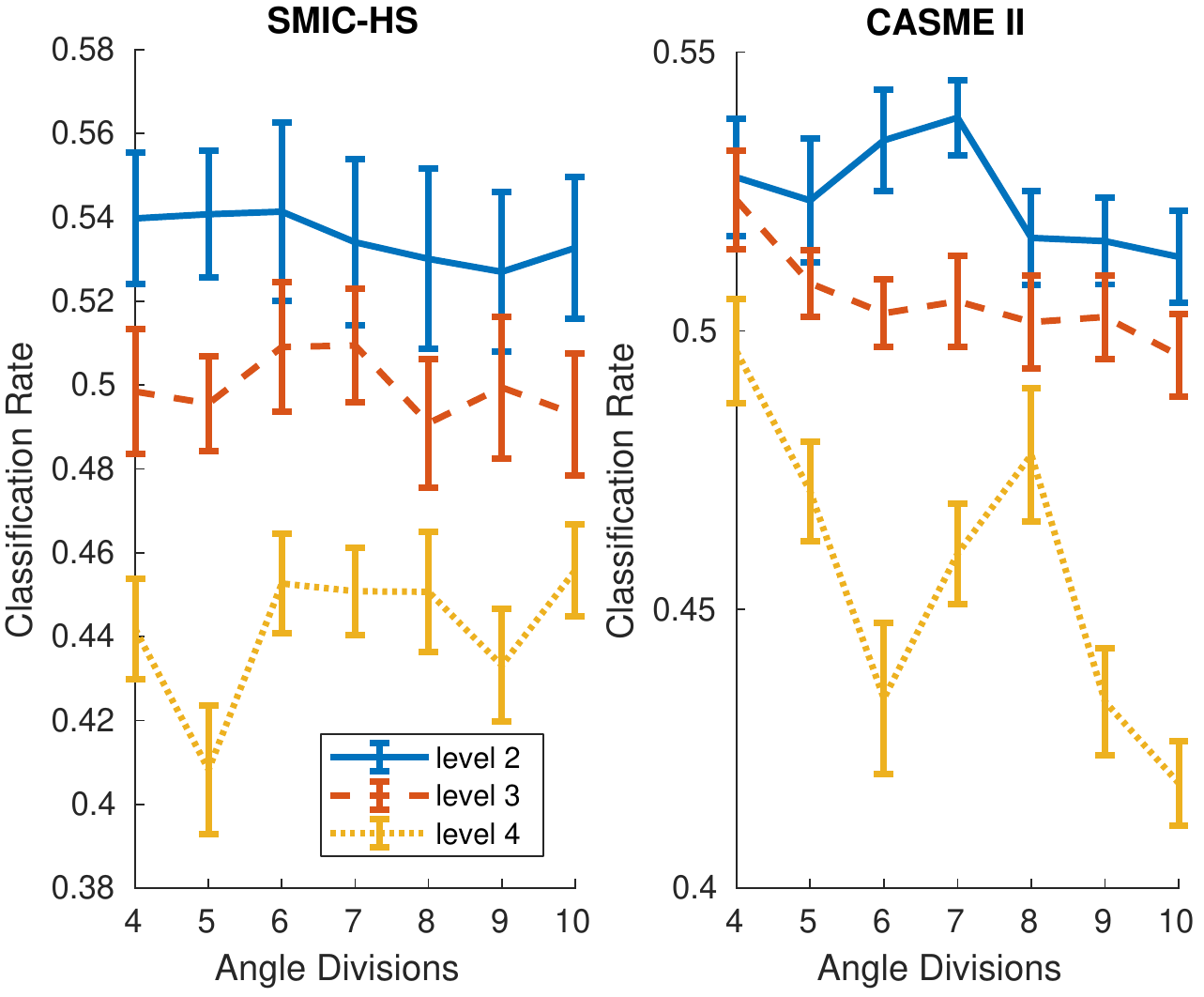}
    \caption{Orientation binning parameter evaluation}
    \label{fig:paraangle}
\end{figure}

\section{Experimental Results}
\label{sec:exp}
\subsection{Datasets}
For our experimentation, two spontaneously elicited ME databases are selected. First, the SMIC database~\citep{li_spontaneous_2013} consists of 164 spontaneous facial MEs image sequences from 16 subjects. The full version of SMIC contains three datasets: the SMIC-HS dataset recorded by a high speed camera at 100 fps; the SMIC-VIS dataset recorded by a color camera at 25 fps; and the SMIC-NIR dataset recorded by a near infrared camera at 25 fps (all with a spatial resolution of $640\times 480$). Ground truth annotations provide the frame numbers indicating the onset and offset frames. The MEs are labeled into three emotion classes: positive, surprise and negative emotions. For our experimentation it is decided to use only the SMIC-HS dataset.

Secondly, the CASME II ~\citep{yan_casme_2014} database consists of 247 spontaneous facial MEs image sequences from 26 subjects. They were recorded using a high speed camera at 200 fps and spatial resolution of $640\times 480$. Ground truth annotations not only provide the frame numbers indicating the onset and offset but also the apex frames (the moment when the ME is at its highest intensity). The MEs are labeled into five classes: happiness, surprise, disgust, repression and others.

\begin{figure}[H]
    \centering
    \subfloat[SMIC-HS\label{fig:paragridSMIC}]{\includegraphics[width=0.49\columnwidth]{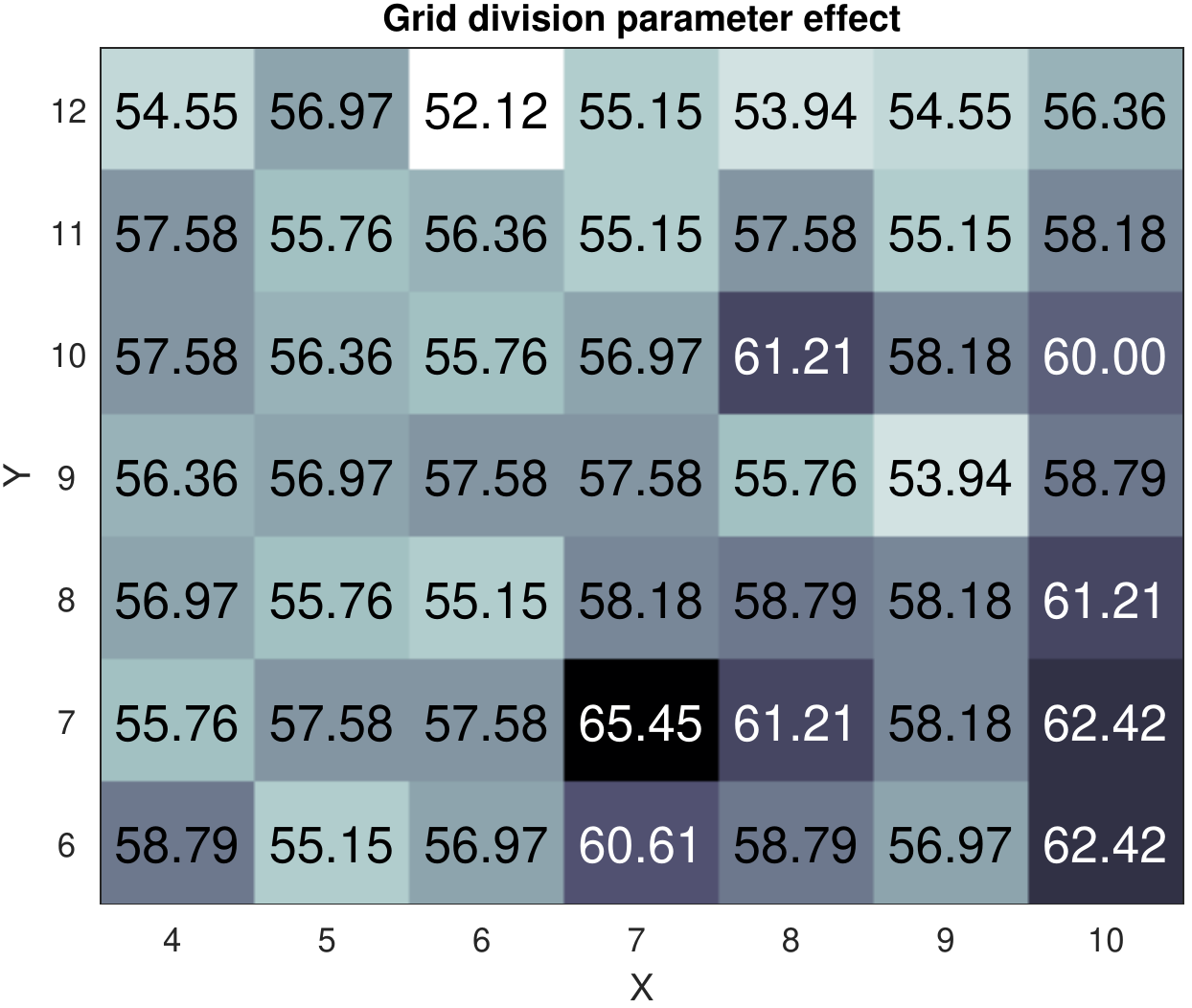}}
    \subfloat[CASME II\label{fig:paragridCASME}]{\includegraphics[width=0.49\columnwidth]{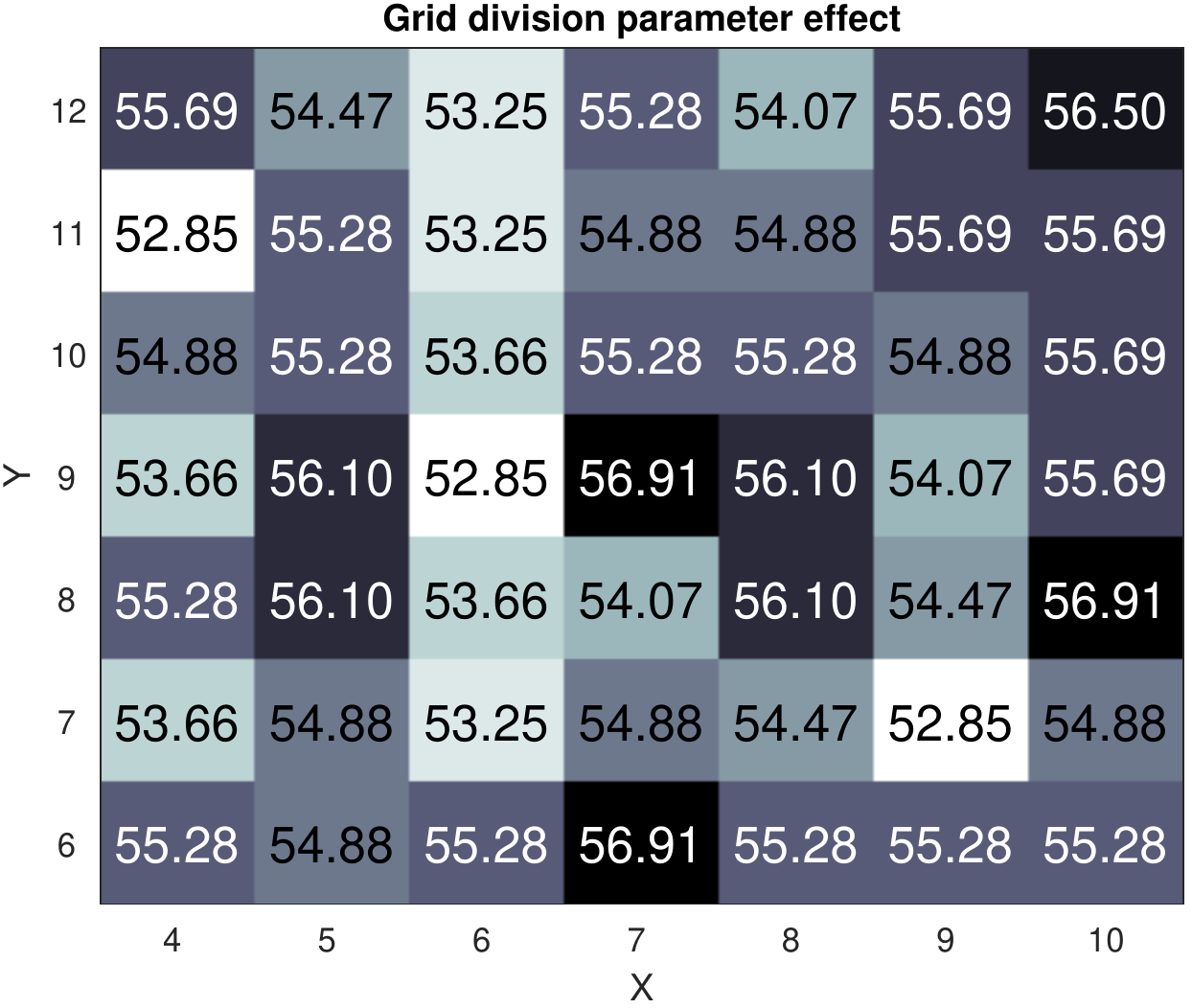}}
    \caption{Grid division parameter evaluation}
    \label{fig:paragrid}
\end{figure}

\subsection{Parameter Analysis}
To evaluate the impact of the different parameters on our system, our proposed framework is tested while its parameter values are varied. It is decided to evaluate the following parameters: the pyramid level (from 2nd to 4th level\footnote{The first level has the information of the highest frequency sub-band and seems to carry an important amount of undesired noise, therefore is not included in our analysis.}), the grid division ($G_x=[4,\dots,10]$ and $G_y=[6,\dots,12]$) and orientation binning ($O=[4,\dots,10]$). For classification, a multiclass-SVM (one-vs-all) using a polynomial kernel of degree three is trained. The hyperparameters are tuned by an exhaustive grid search process. The results are tested by a Leave-One-Subject-Out (LOSO) cross-validation. 

The effect of the level of the Riesz pyramid is shown (Fig.~\ref{fig:paralev}). In both the SMIC-HS dataset and the CASME II dataset extracting the phase values from the 2nd level of the pyramid results in a higher median accuracy. Furthermore, the effect of the angle division for the MORF descriptor is shown (Fig.~\ref{fig:paraangle}). In both cases, the best angle division for the level that yields the best results (2nd) is between 6 and 7 divisions. The effect of the grid division for the MORF descriptor is also shown (Fig.~\ref{fig:paragrid}). In the SMIC-HS dataset, the best grid division is between 7 and 10 for $G_x$ and 6 to 8 for $G_y$. In the CASME II dataset, the best grid division is between 7 and 9 for $G_x$ and 8 to 10 for $G_y$ (However there are some peaks when $G_x$ is 10).

\subsection{MORF variations}
Some variations to the MORF descriptor are tested by combining different data and methodologies. Firstly, the results of two or more MORF histograms are merged from different levels of the Riesz pyramid (\textbf{F-MORF}). The idea is to use the oriented phase calculated from different sub-bands to potentially complement the information for modelling an ME. Secondly, it is decided to use the amplification process of \citep{wadhwa_phase-based_2013} in which the subtle phase changes are multiplied by a scalar without amplifying the noise (\textbf{A-MORF}). This is done by multipling the quaternionic filtered phase $(\phi\cos(\theta),\phi\sin(\theta))$ by a magnification factor ($\alpha$), then, after performing a quaternion exponentiation on it, the amplified quaternioninc phase is extracted:
    \begin{equation}
        \sin(\alpha\phi)\cos(\theta),\sin(\alpha\phi)\sin(\theta)
    \label{eq:ampquatphase}
\end{equation}
Finally, this representation is used to calculate the MOR image pair and extract the MORF descriptor. The pyramid levels can also be both merged and amplified to obtain \textbf{AF-MORF}. The recognition performance of the proposed method is measured using both recognition accuracy and F-measure. The results are shown in Table~\ref{tab:ex3smic} and  Table~\ref{tab:ex3casme}.

\begin{table}[!t]
    \centering
    \small
    \resizebox{\columnwidth}{!}{%
    \begin{tabular}{|c|c|c|c|c|c|c|c|}
    \hline
    \multicolumn{8}{|c|}{SMIC-HS}  \\ \hline
    \multirow{3}{*}{Feature} & \multirow{3}{*}{\shortstack{Pyramid \\ Level}} & \multicolumn{2}{c|}{\multirow{2}{*}{\shortstack{Non \\ Amplified}}} & \multicolumn{4}{c|}{A-MORF} \\ \cline{5-8} 
    & & \multicolumn{2}{c|}{} & \multicolumn{2}{c|}{$\alpha = 5$} & \multicolumn{2}{c|}{$\alpha = 10$} \\ \cline{3-8} 
                        &               & Acc          & F1-mea     & Acc            & F1-mea        & Acc             & F1-mea      \\ \hline
    \multirow{3}{*}{MORF}   & $2$       & $\mathbf{65.45\%}$    & $\mathbf{0.6466}$   & $58.79\%$      & $0.5764$      & $53.33\%$       & $0.5266$    \\ \cline{2-8} 
                            & $3$       & $57.58\%$    & $0.5733$   & $61.21\%$      & $0.6059$      & $58.18\%$       & $0.5822$    \\ \cline{2-8} 
                            & $4$       & $50.30\%$    & $0.5043$   & $50.30\%$      & $0.5054$      & $50.91\%$       & $0.5158$    \\ \hline
    \multirow{3}{*}{F-MORF} & $2\&3$    & $58.79\%$    & $0.5937$   & $60.00\%$      & $0.5999$      & $59.39\%$       & $0.5895$    \\ \cline{2-8} 
                            & $3\&4$    & $54.55\%$    & $0.5507$   & $56.36\%$      & $0.5695$      & $58.18\%$       & $0.5798$    \\ \cline{2-8} 
                            & $2\&3\&4$ & $58.79\%$    & $0.5913$   & $58.79\%$      & $0.5794$      & $59.39\%$       & $0.5868$    \\ \hline
    \end{tabular}}
    \caption{ME classification for SMIC HS in terms of accuracy and F-measure}
	\label{tab:ex3smic}
\end{table}

\begin{table}[!t]
    \centering
    \small
    \resizebox{\columnwidth}{!}{%
    \begin{tabular}{|c|c|c|c|c|c|c|c|}
    \hline
    \multicolumn{8}{|c|}{CASME II}  \\ \hline
    \multirow{3}{*}{Feature} & \multirow{3}{*}{\shortstack{Pyramid \\ Level}} & \multicolumn{2}{c|}{\multirow{2}{*}{\shortstack{Non \\ Amplified}}} & \multicolumn{4}{c|}{A-MORF} \\ \cline{5-8} 
    & & \multicolumn{2}{c|}{} & \multicolumn{2}{c|}{$\alpha = 5$} & \multicolumn{2}{c|}{$\alpha = 10$} \\ \cline{3-8} 
                            &           & Acc          & F1-mea     & Acc            & F1-mea        & Acc             & F1-mea      \\ \hline
    \multirow{3}{*}{MORF}   & $2$       & $56.91\%$    & $0.5878$   & $58.94\%$      & $0.6045$      & $58.54\%$       & $0.5983$    \\ \cline{2-8} 
                            & $3$       & $55.69\%$    & $0.5545$   & $57.72\%$      & $0.5779$      & $58.94\%$       & $0.5762$    \\ \cline{2-8} 
                            & $4$       & $52.85\%$    & $0.5110$   & $53.66\%$      & $0.5183$      & $54.88\%$       & $0.5355$    \\ \hline
    \multirow{3}{*}{F-MORF} & $2\&3$    & $58.54\%$    & $0.5923$   & $\mathbf{62.20\%}$      & $\mathbf{0.6304}$      & $62.20\%$       & $0.6171$    \\ \cline{2-8} 
                            & $3\&4$    & $59.35\%$    & $0.5871$   & $56.10\%$      & $0.5549$      & $56.50\%$       & $0.5576$    \\ \cline{2-8} 
                            & $2\&3\&4$ & $59.76\%$    & $0.6012$   & $58.54\%$      & $0.5850$      & $58.13\%$       & $0.5827$    \\ \hline
    \end{tabular}}
    \caption{ME classification for CASME II in terms of accuracy and F-measure}
	\label{tab:ex3casme}
\end{table}

For SMIC-HS better results are obtained using MORF and CASME II using FA-MORF. This discrepancy comes from the differences of the datasets. The subjects in CASME II were at a closer distance to the camera during video recording compared to SMIC, thus the captured faces had a bigger resolution which result in a shift of the ME motion to low frequencies. This might explain why better results can be obtained using the 2nd level of the Riesz pyramid in the SMIC-HS dataset but in the CASME II dataset similar results are obtained both in the 2nd and 3rd levels. Consequently merging these two levels yields better results for CASME II but not for SMIC-HS. Furthermore the CASME II videos were captured with a camera twice as fast as those in SMIC-HS. This means that the phase differences between frames in the CASME II database are smaller and can potentially be improved by amplification which might explain why A-MORF and FA-MORF perform better in the CASME II dataset.

\subsection{State-of-the-art Comparison}
Our classification results are compared with some representative methods from the state of the art\footnote{For a more thorough comparison, we refer the reader to state-of-the-art surveys presented in \citep{oh_survey_2018} and \citep{goh_micro-expression_2018}} in Tab.\ref{tab:ressoa}. For the LBP-based methods, it can be seen how they have improved from the baseline proposed by \citep{li_spontaneous_2013}. Each method extracts spatio-temporal information by creating a different code-book based on the intensity difference between a pixel and its 3D neighborhood. One reason why these methods tend to do better in the dataset CASME II is that, as previously mentioned, its images have a bigger resolution which means extracting better textured information of the MEs. For the OF-based methods, each method extracts motion information from OF. The best results come from Bi-WOOF \citep{liong_optical_2014} and OF Maps \citep{allaert_consistent_2017} for calculating the motion between onset and apex frames (instead of calculating the motion between consecutive frames). Furthermore, OF Maps extract the coherent movement on the face in different locations and use it to filter residual motion vectors (noise). For the deep learning methods, each method either trains or tunes a pre-trained convolutional neural network (CNN) for extracting features and classification. Although these methods are becoming more widely used in classification problems, they also struggle to obtain good results when dealing with small datasets. And although they obtain good results with the CASME II, they avoid using a smaller dataset such as SMIC-HS.

Our proposed approach does not yield the best possible results. It is worth noting that, while other authors have been able to obtain better results starting from a baseline method (like LBP-TOP in the case of the LBP-based methods and HOOF in the case of OF-based methods) and propose an improved method by changing the way they extract features  and how to code them (quantizing the values, developing a weighted histogram, pre-selecting regions of interest and/or frames to process, etc.), our method uses a more basic featuring extracting method. It can be imagined that applying a more sophisticated method to extract information from Riesz phase variations might result in better results. All things considered, our method is still able to surpass several descriptors in both datasets. 

Our method outperforms other Riesz based methods like \citep{oh_monogenic_2015,oh_intrinsic_2016} by approximately $20\%$. However, \citep{liong_micro-expression_2017} remains as the best Riesz phase-based method. It's worth noting that the performance of the previous version of this method, Bi-WOOF \citep{liong_optical_2014}, is greatly improved by adding Riesz phase difference between onset and apex frames \citep{liong_micro-expression_2017}. This implies that our results could be improved by complementing our phase-based features with other motion and texture features.  

One cause of error might come from the limitations of the local orientation of the monogenic signal. The local orientation $\theta$ represents the dominant direction in the image at any given point. This representation comes from the formulation of the monogenic signal which assumes images as intrinsically one-dimensional signals. This means that the monogenic signal is useful for modelling image features such as edges and lines that have variation in one direction only, but cannot model image features such as corners that have variation in two directions (intrinsically 2D signals)~\citep{wietzke_geometry_2009}.

\begin{savenotes}
\begin{table*}
    \centering
    \small
    \begin{tabular}{|x{1.1cm}|x{3.5cm}|x{3.4cm}|x{1.4cm}|x{1.4cm}|x{1.4cm}|x{1.4cm}|}
        \hline
        \multicolumn{7}{|c|}{Micro-Expression Classification Methods} \\ 
        \hline
        \multirow{2}{*}{Family} & \multicolumn{2}{c|}{Method} & \multicolumn{2}{c|}{Accuracy} & \multicolumn{2}{c|}{F-measure} \\\cline{2-7}
        & Features & Paper & SMIC HS & CASME II & SMIC HS & CASME II \\
		%%%%%%%%%%%%%%%%%%%%%%%%%%%%%%%%%%%%%%%%%%%%%%%%%%%%%%%%%%%%%%%%%%%%%
		\hline
	    \multirow{5}{*}{\shortstack{LBP \\ based}}
    	& LBP-TOP & \cite{li_spontaneous_2013}        & $48.78\%$ & $-$ & $-$ & $-$ \\ 
        & LBP-MOP & \cite{wang_efficient_2015}        & $50.61\%$ & $45.75\%$ & $-$ & $-$\\
        & STLBP-IP & \cite{huang_facial_nodate}       & $57.93\%$ & $59.51\%$ & $0.58$\footnote[3]{Values extracted from \cite{oh_survey_2018}} & $0.57$\footnotemark[3] \\
        & STCQLP & \cite{huang_spontaneous_2016-1}    & $58.39\%$ & $64.02\%$ & $0.6381$ & $0.5836$ \\
		& Di-STLBP-IP & \cite{huang_spontaneous_2016} & $63.41\%$ & $64.78\%$ & $-$ & $-$ \\
	   % & & \cite{wang_micro-expression_2014}& $67.68\%$ & $65.44\%$\\
         %%%%%%%%%%%%%%%%%%%%%%%%%%%%%%%%%%%%%%%%%%%%%%%%%%%%%%%%%%%%%%%%%%%%%
        \hline
        \multirow{5}{*}{\shortstack{OF \\ based}} 
        % & \citep{liu_main_2016}              & $58.97\%$ & $51.69\%$ \\
        & OS & \cite{liong_optical_2014}              & $53.56\%$ & $-$ & $-$ & $-$ \\
	    & FDM & \cite{xu_microexpression_2016}        & $54.88\%$ & $41.96\%$ & $0.538$ & $0.4053$ \\
        & HFOFO & \cite{happy_fuzzy_2018}             & $51.83\%$ & $56.64\%$ & $0.5243$ & $0.5248$\\
        & Bi-WOOF & \cite{liong_optical_2014}         & $62.20\%$ & $58.85\%$ & $0.62$ & $0.61$\\
        & OF Maps & \cite{allaert_consistent_2017}    & $-$ & $\mathbf{65.35}\%$ & $-$ & $-$ \\
        % & & \cite{lu_motion_2018}             & $71.95\%$ & $69.11\%$ \\
	     %%%%%%%%%%%%%%%%%%%%%%%%%%%%%%%%%%%%%%%%%%%%%%%%%%%%%%%%%%%%%%%%%%%%%
        \hline
        \multirow{4}{*}{\shortstack{Deep \\ learning}} 
        % & \citep{he_multi-task_2017}     & $63.95\%$ & $-$   \\
        & Imagenet & \cite{patel_selective_2016}      & $53.60\%$ & $47.30\%$ & $-$ & $-$ \\
        & 3D-FCNN & \cite{li_micro-expression_2018}   & $55.49\%$ & $59.11\%$ & $-$ & $-$ \\
        & CNN + LSTM & \cite{kim_micro-expression_2016}     & $-$ & $60.98\%$ & $-$ & $-$ \\
        & VGGNet & \cite{li_can_2018}                       & $-$ & $63.30\%$ & $-$ & $-$ \\
        %%%%%%%%%%%%%%%%%%%%%%%%%%%%%%%%%%%%%%%%%%%%%%%%%%%%%%%%%%%%%%%%%%%%%
        \hline
        \multirow{4}{*}{Others}
        & Monogenic + LBP-TOP & \cite{oh_monogenic_2015} & $-$ & $-$ & $0.44$ & $0.41$ \\
        & Riesz Wavelet + LBP-TOP & \cite{oh_intrinsic_2016} & $-$ & $-$ & $-$ & $0.43$ \\
        & OSW-LBP-TOP & \cite{liong_subtle_2014}      & $53.66\%$ & $42.00\%$ & $0.54$ & $0.38$ \\
        & ST-Gabor & \cite{lin_micro-expression_2018} & $54.47\%$ & $55.28\%$ & $-$ & $-$ \\ 
        & Bi-WOOF + Riesz Phase & \cite{liong_micro-expression_2017}                                                                               & $\mathbf{68.29}\%$ & $62.55\%$ & $\mathbf{0.67}$ & $\mathbf{0.65}$ \\
        %%%%%%%%%%%%%%%%%%%%%%%%%%%%%%%%%%%%%%%%%%%%%%%%%%%%%%%%%%%%%%%%%%%%%
        \hline
        \multirow{3}{*}{\shortstack{Our \\ method}} 
        & \multicolumn{2}{c|}{MORF}    & $\mathbf{65.45}\%$ & $56.91\%$ & $\mathbf{0.6466}$ & $0.5878$\\ 
        & \multicolumn{2}{c|}{F-MORF}  & $58.79\%$ & $59.76\%$ & $0.5937$ & $0.6012$ \\ 
        & \multicolumn{2}{c|}{FA-MORF} & $61.21\%$ & $\mathbf{62.20}\%$ & $0.6059$ & $\mathbf{0.6304}$\\ 
		\hline
    \end{tabular}
	\caption{Comparison of micro-expression recognition performance in terms of accuracy and F-measure for feature-extraction state-of-the-art methods.}
	\label{tab:ressoa}
\end{table*}
\end{savenotes}

\section{Conclusion}
\label{sec:conc}
A facial micro-expression recognition method based on the quaternionic oriented phase representation of the multi-scale monogenic signals is proposed. Phase variations are quickly extracted from a video using an approximate Riesz transform called the Riesz pyramid.
The temporal evolution of a micro-expression is modelled as an image pair that contains the mean oriented phase component of the monogenic signal which aims to reduce the effects of image noise. Furthermore, this model is extended into an easily-adaptable and low-dimensional feature descriptor which can also contain the amplification of the oriented phase or concatenate the multi-scale oriented phase representation. 
Our method achieves an accuracy of $65.45\%$ and an F-score of $0.6466$ for the SMIC-HS dataset and an accuracy of $62.20\%$ and an F-score of $0.6304$ for the CASME II dataset. These are the best results for methods focused on Riesz transform based features. It is also competitive against methods based on widely researched features such as LBP and OF and even against deep learning methods. Furthermore, it obtains the best results in the SMIC-HS for methods focused in one single type of feature.
% Experiments also show that our method is robust with changes in parameters and some preliminary results are given showing that the proposed descriptor is promising. 
% Our method has shown to be a powerful tool for ME recognition, comparable to other state of the art methods based on classical features.
% The Riesz pyramid has shown itself to be a powerful tool for ME recognition comparable to other methods of the state of the art that follow more classic basis.
% Our method has shown to be a powerful tool for ME recognition

Our Riesz pyramid-based method has shown itself to be a powerful tool for ME recognition.
Adopting a more sophisticated feature extraction and codification, along with complementing the Riesz phase variations with motion or texture information, could be used to create an improved ME analysis technique in the future. In addition, since there is already a method that uses a similar basis for ME spotting \cite{8354118}, both methods can be merged for an integrated Riesz phase-based spotting and recognition framework.

\bibliographystyle{model2-names}
\bibliography{CarlosPhdPeople2}

\end{document}